\documentclass[11pt]{article}
\usepackage{acl2013}
\usepackage{times}
\usepackage{url}
\usepackage{latexsym}
\usepackage{graphicx}
\usepackage{algorithm}
\usepackage[noend]{algpseudocode}
\usepackage{multirow}

\usepackage{amsmath}
\usepackage{amssymb}
\usepackage{bm}
\usepackage{balance}

\title{Answer Sequence Learning with Neural Networks for Answer Selection in Community Question Answering}

\author{ Xiaoqiang Zhou~~~~~ Baotian Hu~~~~~ Qingcai Chen\thanks{* Corresponding author}~~~~~ Buzhou Tang~~~~~ Xiaolong Wang
\\
\\
\begin{tabular}{c}
 { Intelligence Computing Research Center}    \\
 {Harbin Institute of Technology, Shenzhen Graduate School}            \\
 {\tt \{xiaoqiang.jeseph,baotianchina,qingcai.chen,tangbuzhou\}@gmail.com} \\
 {\tt wangxl@insun.hit.edu.cn} \\
\end{tabular}
}

\begin{document}
\maketitle
\begin{abstract}

In this paper, the answer selection problem in community question answering (CQA) is regarded as an answer sequence labeling task, and a novel approach is proposed based on the recurrent architecture for this problem. Our approach applies convolution neural networks (CNNs) to learning the joint representation of question-answer pair firstly, and then uses the joint representation as input of the long short-term memory (LSTM) to learn the answer sequence of a question for labeling the matching quality of each answer. Experiments conducted on the SemEval 2015 CQA dataset shows the effectiveness of our approach.

\end{abstract}

\section{Introduction}

\begin{figure}[!tb]
\centering
\includegraphics[width=.48\textwidth]{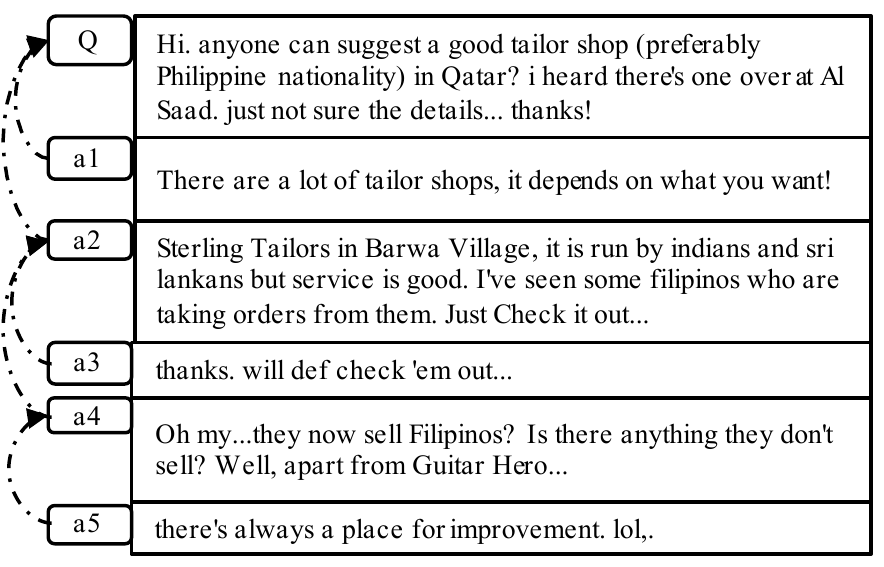}
\caption{An Example of the Answer Sequence for a Question. The dashed arrows depict the relationships of the answers in the sequence.}
\label{aclexample}
\vspace{-4pt}
\end{figure}

Answer selection in community question answering (CQA), which recognizes high-quality responses to obtain useful question-answer pairs, is greatly valuable for knowledge base construction and information retrieval systems. To recognize matching answers for a question, typical approaches model semantic matching between question and answer by exploring various features~\cite{baoxun2009,Shah:2010}. Some studies exploit syntactic tree structures~\cite{wang:2009,alessandro2007} to measure the semantic matching between question and answer. However, these approaches require high-quality data and various external resources which may be quite difficult to obtain. To take advantage of a large quantity of raw data, deep learning based approaches ~\cite{baoxun2010,hu:2013} are proposed to learn the distributed representation of question-answer pair directly. One disadvantage of these approaches lies in that semantic correlations embedded in the answer sequence of a question are ignored, while they are very important for answer selection. 
Figure~\ref{aclexample} is a example to show the relationship of  answers in the sequence for a given question. Intuitively, other answers of the question are beneficial to judge the quality of the current answer. 
 
Recently, recurrent neural network (RNN), especially Long Short-Term Memory (LSTM)~\cite{hochreiter}, has been proved superiority in various tasks~\cite{seq2seq,lstmuse} and it models long term and short term information of the sequence. And also, there are some works on using convolutional neural networks (CNNs) to learn the representations of sentence or short text, which achieve state-of-the-art performance on sentiment classification~\cite{sentiment} and short text matching ~\cite{baotian}.

In this paper, we address the answer selection problem as a sequence labeling task, which identifies the matching quality of each answer in the answer sequence of a question. Firstly, CNNs are used to learn the joint representation of question answer (QA) pair. Then the learnt joint representations are used as inputs of LSTM to predict the quality (e.g., {\em Good}, {\em Bad} and {\em Potential}) of each answer in the answer sequence. Experiments conducted on the CQA dataset of the answer selection task in SemEval-2015\footnote{http://alt.qcri.org/semeval2015/task3/} show that the proposed approach outperforms other state-of-the-art approaches.

\section{Related Work}
Prior studies on answer selection generally treated this challenge as a classification problem via employing machine learning methods, which rely on exploring various features to represent QA pair. ~\newcite{huang:2007} integrated textual features with structural features of forum threads to represent the candidate QA pairs, and used support vector machine (SVM) to classify the candidate pairs. Beyond typical features, ~\newcite{Shah:2010} trained a logistic regression (LR) classifier with user metadata to predict the quality of answers in CQA. ~\newcite{ding2008} proposed an approach based on conditional random fields (CRF), which can capture contextual features from the answer sequence for the semantic matching between question and answer. Additionally, the translation-based language model was also used for QA matching by transferring the answer to the corresponding question~\cite{jeon:2005,xuetrans,zhou:2011}. The translation-based methods suffer from the informal words or phrases in Q\&A archives, and perform less applicability in new domains.

In contrast to symbolic representation, ~\newcite{baoxun2010} proposed a deep belief nets (DBN) based semantic relevance model to learn the distributed representation of QA pair. Recently, the convolutional neural networks (CNNs) based sentence representation models have achieved successes in neural language processing (NLP) tasks. ~\newcite{dl2qa} proposed a convolutional sentence model to identify answer contents of a question from Q\&A archives via means of distributed representations. The work in ~\newcite{baotian} demonstrated that 2-dimensional convolutional sentence models can represent the hierarchical structures of sentences and capture rich matching patterns between two language objects.
\section{Approach}
We consider the answer selection problem in CQA as a sequence labeling task. To label the matching quality of each answer for a given question, our approach models the semantic links between successive answers, as well as the semantic relevance between question and answer. Figure 1 summarizes the recurrent architecture of our model (R-CNN). The motivation of R-CNN is to learn the useful context to improve the performance of answer selection. The answer sequence is modeled to enrich semantic features.

\begin{figure}[!tb]
\centering
\includegraphics[width=.5\textwidth]{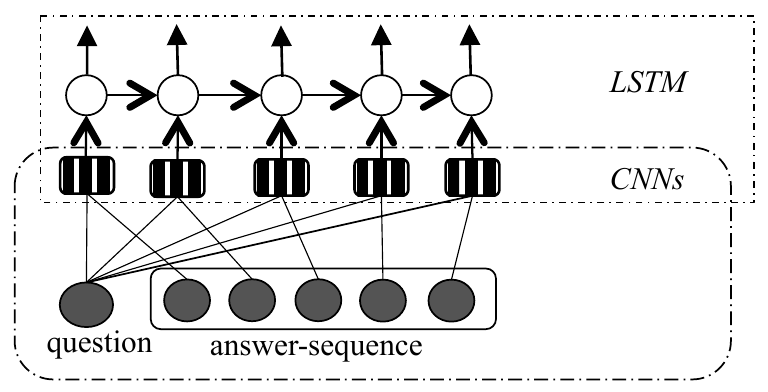}
\caption{The architecture of R-CNN}
\label{figure1}
\end{figure}

At each step, our approach uses the pre-trained word embeddings to encode the sentences of QA pair, which then is used as the input vectors of the model. Based on the joint representation of QA pair learned from CNNs, the LSTM is applied in our model for answer sequence learning, which makes a prediction to each answer of the question with softmax function.

\subsection{Convolutional Neural Networks for QA Joint Learning}
Given a question-answer pair at the step $t$, we use convolutional neural networks (CNNs) to learn the joint representation $p_t$ for the pair. Figure 2 illustrates the process of QA joint learning, which includes two stages: summarizing the meaning of the question and an answer, and generating the joint representation of QA pair.

To obtain high-level sentence representations of the question and answer, we set 3 hidden layers in two convolutional sentence models respectively. The output of each hidden layer is made up of a set of 2-dimensional arrays called feature map parameters $(w_m,b_m )$. Each feature map is the outcome of one convolutional or pooling filter. Each pooling layer is followed an activation function $\sigma$. The output of the $m^{th}$ hidden layer is computed as Eq.~\ref{eq1}:
\begin{equation}
H_m = \sigma(pool(w_mH_{m-1}+b_m))
\label{eq1}
\end{equation}
\noindent Here, $H_0$ is one real-value matrix after sentence semantic encoding by concatenating the word vectors with sliding windows. It is the input of deep convolution and pooling, which is similar to that of traditional image input.

Finally, we combine the two sentence models by adding an additional layer $H_t$  on the top. The learned joint representation $p_t$ for QA pair is formalized as Eq.~\ref{eq2}:
\begin{equation}
p_t=\sigma(w_t H_t+b_t )   
\label{eq2}
\end{equation}
\noindent where $\sigma$ is an activation function, and the input vector is constructed by concatenating the sentence representations of question and answer.

\begin{figure}[!tb]
\centering
\includegraphics[width=.5\textwidth]{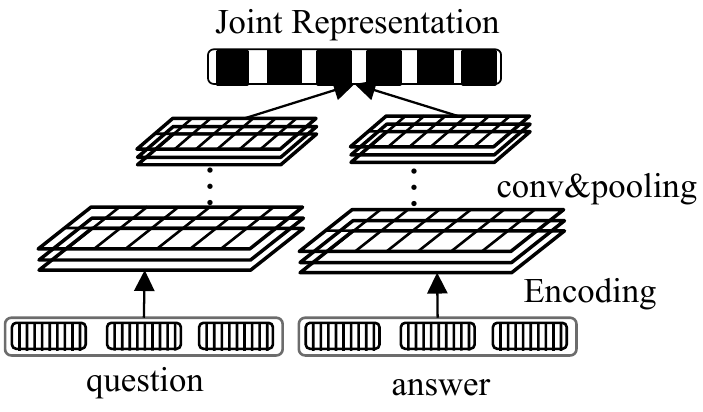}
\caption{CNNs for QA joint learning}
\label{figure2}
\end{figure}

\subsection{LSTM for Answer Sequence Learning}

Based on the joint representation of QA pair, the LSTM unit of our model performs answer sequence learning to model semantic links between continuous answers. Unlike the traditional recurrent unit, the LSTM unit modulates the memory at each time step, instead of overwriting the states. The key component of LSTM unit is the memory cell $c_t$ which has a state over time, and the LSTM unit decides to modify and add the memory in the cell via the sigmoidal gates: input gate $i_t$, forget gate $f_t$ and output gate $o_t$. The implementation of the LSTM unit in our study is close the one discussed by~\newcite{graves13}.
Given the joint representation $p_t$ at time $t$, the memory cell $c_{t}$ is updated by the input gate's activation $i_t$ and the forget gate's activation $f_t$. The updating equation is given by Eq.~\ref{eq3}:
\begin{equation}
c_t=f_tc_{t-1}+i_ttanh(W_{xc}p_t+W_{hc}h_{t-1}+b_c)   
\label{eq3}
\end{equation}

The LSTM unit keeps to update the context by discarding the useless context in forget gate $f_t$ and adding new content from input gate $i_t$. The extents to modulate context for these two gates are computed as Eq.~\ref{eq4} and Eq.~\ref{eq5}:

\begin{equation}
i_t=\sigma(W_{xi}p_t+W_{hi}h_{(t-1)}+W_{ci}c_{t-1}+b_i )
\label{eq4}
\end{equation}

\begin{equation}
f_t=\sigma(W_{xf}p_t+W_{hf}h_{t-1}+W_{cf}c_{t-1}+b_f )
\label{eq5}
\end{equation}

With the updated cell state $c_t$, the final output from LSTM unit $h_t$  is computed as Eq~\ref{eq6}:
\begin{equation}
o_t=\sigma(W_{xo}p_t+W_{ho}h_{t-1}+W_{co}c_t+b_o )          
\label{eq6}
\end{equation}

\begin{equation}
h_t=o_t tanh(c_t)                  
\label{eq7}
\end{equation}

Note that $(W_*,b_*)$ is the parameters of LSTM unit, in which $W_{cf}, W_{ci}$ , and $W_{co}$ are diagonal matrices.

According to the output $h_t$ at each time step, our approach estimates the conditional probability of the answer sequence over answer classes, it is given by Eq.~\ref{eq8}:
\begin{equation}
\begin{aligned}
 P(y_1,...,y_T|c,p_1,...,p_{t-1})&=\\
 \prod_{t=1}^Tp(y_t|c,y_1,...,y_{t-1}) &    
\end{aligned}
\label{eq8}
\end{equation}

\noindent Here, $(y_1,...,y_T)$ is the corresponding label sequence for the input sequence $(p_1,...,p_{t-1})$, and the class distribution $p(y_t|c,y_1,...,.y_{t-1})$ is represented by a softmax function.

\begin{table}
\centering
\begin{tabular}{llll}
    \hline
    Data      & \#question &  \#answer &  length\\
    \hline
    training       &  2600  &  16541 &  6.36\\
   development          &   300 & 1645 & 5.48\\
   test        &   329 & 1976 & 6.00\\   
   \hline
   all       &   3229 & 21062 & 6.00\\       \hline
\end{tabular}
\caption{Statistics of experimental dataset}
\label{table1}
\vspace{-5pt}
\end{table}

\section{Experiments}
\label{section-experiments}
\subsection{Experiment Setup}

\textbf{Experimental Dataset:} We conduct experiments on the public dataset of the answer selection challenge in SemEval 2015. This dataset consists of three subsets: training, development, and test sets, and contains 3,229 questions with 21,062 answers. The answers falls into three classes: {\em Good}, {\em Bad}, and {\em Potential}, accounting for 51\%, 39\%, and 10\% respectively. The statistics of the dataset are summarized in Table~\ref{table1}, where ¡°\#question/answer¡± denotes the number of questions/answers, and ¡°length¡± stands for the average number of answers for a question.

\noindent \textbf{Competitor Methods:} We compare our approach against the following competitor methods:

SVM~\cite{huang:2007}: An SVM-based method with bag-of-words (textual features), non-textual features, and features based on topic model (i.e., latent Dirichlet allocation, LDA).

CRF~\cite{ding2008}: A CRF-based method using the same features as the SVM approach.

DBN~\cite{baoxun2010}: Taking bag-of-words representation, the method applies deep belief nets to learning the distributed representation of QA pair, and predicts the class of answers using a logistic regression classifier on the top layer.

mDBN~\cite{hu:2013}: In contrast to DBN, multimodal DBN learns the joint representations of textual features and non-textual features rather than bag-of-words.

CNN: Using word embedding, the CNNs based model in ~\newcite{baotian} is used to learn the representations of questions and answers, and a logistic regression classifier is used to predict the class of answers.

\noindent \textbf{Evaluation Metrics:} The evaluation metrics include $Macro-precision (P)$, $Macro-recall (R)$, $Macro-F1 (F1)$, and $F1$ scores of the individual classes. According to the evaluation results on the development set, all the hyperparameters are optimized on the training set.

\noindent \textbf{Model Architecture and Training Details:} The CNNs of our model for QA joint representation learning have 3 hidden layers for modeling question and answer sentence respectively, in which each layer has 100 feature maps for convolution and pooling operators. The window sizes of convolution for each layer are $[1\times1,2\times2,2\times2]$, the window sizes of pooling are $[2\times2,2\times2,1\times1]$. For the LSTM unit, the size of {\em input gate} is set to 200, the sizes of {\em forget gate}, {\em output gate}, and {\em memory cell} are all set to 360.

Stochastic gradient descent (SGD) algorithm via back-propagation through time is used to train the model. To prevent serious overfitting, early stopping and dropout~\cite{hinton2012} are used during the training procedure. The learning rate $\lambda$ is initialized to be 0.01 and is updated dynamically according to the gradient descent using the ADADELTA method~\cite{adagrad}. The activation functions $(\sigma,\gamma)$ in our model adopt the rectified linear unit (ReLU)~\cite{relu}.  In addition, the word embeddings for encoding sentences are pre-trained with the unsupervised neural language model~\cite{wordvec} on the Qatar Living data\footnote{http://alt.qcri.org/semeval2015/task3/index.php?id=data-and-tools}.

\subsection{Results and Analysis}

Table~\ref{table2} summarizes the Macro-averaged results. The F1 scores of the individual classes are presented in Table~\ref{table3}.

It is clear to see that the proposed R-CNN approach outperforms the competitor methods over the Macro-averaged metrics as expected from Table~\ref{table2}. The main reason lies in that R-CNN takes advantages of the semantic correlations between successive answers by LSTM, in addition to the semantic relationships between question and answer. The joint representation of QA pair learnt by CNNs also captures richer matching patterns between question and answer than other methods.

It is notable that the methods based on deep learning perform more powerful than SVM and CRF, especially for complicate answers (e.g., {\em Potential} answers). In contrast, SVM and CRF using a large amount of features perform better for the answers that have obvious tendency (e.g., {\em Good} and {\em Bad} answers). The main reason is that the distributed representation learnt from deep learning architecture is able to capture the semantic relationships between question and answer. On the other hand, the feature-engineers in both SVM and CRF suffer from noisy information of CQA and the feature sparse problem for short questions and answers.

\begin{table}
\centering
\begin{tabular}{llll}
    \hline
    Methods      & P&  R &  F1\\
    \hline
    SVM       &  50.10  &  54.43 &  52.14\\
    CRF       &  53.89  &  54.26 &  53.40\\    
    DBN       &  55.22  &  53.80 &  54.07\\    
    mDBN    &  56.11  &  53.95 &  54.29\\  
    CNN    &  55.33  &  54.73 &  54.42\\  
          
   \hline
   R-CNN      &   56.41 & 56.16 & 56.14\\       \hline
\end{tabular}
\caption{Macro-averaged results(\%)}
\label{table2}
\vspace{-5pt}
\end{table}

Compared to DBN and mDBN, CNN and R-CNN show their superiority in modeling QA pair. The convolutional sentence models, used in CNN and R-CNN, can learn the hierarchical structure of language object by deep convolution and pooling operators. In addition, both R-CNN and CNN encode the sentence into one tensor, which makes sure the representation contains more semantic features than the bag-of-words representation in DBN and mDBN.

The improvement achieved by R-CNN over CNN demonstrates that answer sequence learning is able to improve the performance of the answer selection in CQA. Because modeling the answer sequence can enjoy the advantage of the shared representation between successive answers, and complement the classification features with the learnt useful context from previous answers. Furthermore, memory cell and gates in LSTM unit modify the valuable context to pass onwards by updating the state of RNN during the learning procedure.

The main improvement of R-CNN against with the competitor methods comes from the {\em Potential} answers, which are much less than other two type of answers. It demonstrates that R-CNN is able to process the unbalance data. In fact, the {\em Potential} answers are most difficult to identify among the three types of answers as {\em Potential} is an intermediate category~\cite{semval}. Nevertheless, R-CNN achieves the highest F1 score of 15.22\% on Potential answers. In CQA, Q\&A archives usually form one multi-parties conversation when the asker gives feedbacks (e.g., ``ok'' and ``please'') to users¡¯ responses, indicating that the answers of one question are sematic related. Thus, it is easy to understand that R-CNN performs better performance than competitor methods, especially on the recall. The reason is that R-CNN can model semantic correlations between successive answers to learn the context and the long range dependencies in the answer sequence.

\begin{table}
\centering
\begin{tabular}{llll}
    \hline
    Methods      & Good&  Bad &  Potential\\
    \hline
    SVM       &  79.78  &  76.65 &  0.00\\
    CRF       &  79.32  &  75.50 &  5.38\\    
    DBN       &  76.99  &  71.33 &  13.89\\    
    mDBN    &  77.74  &  70.39 &  14.74\\  
    CNN    &  76.45  &  74.77 &  12.05\\  
          
   \hline
   R-CNN      &   77.31 & 75.88 & 15.22\\       \hline
\end{tabular}
\caption{F1 scores for the individual classes(\%)}
\label{table3}
\vspace{-5pt}
\end{table}

\section{Conclusions and Future Work}
In this paper, we propose an answer sequence learning model R-CNN for the answer selection task by integrating LSTM unit and CNNs. Based on the recurrent architecture of our model, our approach is able to model the semantic link between successive answers, in addition to the semantic relevance between question and answer. Experimental results demonstrate that our approach can learn the useful context from the answer sequence to improve the performance of answer selection in CQA.

In the future, we plan to explore the methods on training the unbalance data to improve the overall performances of our approach. Based on this work, more research can be conducted on topic recognition and semantic roles labeling for human-human conversations in real-world.

\paragraph{Acknowledgments:}
{This work was supported in part by National 863 Program of China (2015AA015405), NSFCs (National Natural Science Foundation of China) (61402128, 61473101, 61173075 and 61272383). We thank the anonymous reviewers for their insightful comments.}
\vspace{-10pt}

\bibliographystyle{acl}
\bibliography{acl2015}

\begin{thebibliography}{}

\bibitem[\protect\citename{Dahl \bgroup et al.\egroup }2013]{relu}
George~E. Dahl, Tara~N. Sainath, and Geoffrey~E. Hinton.
\newblock 2013.
\newblock Improving deep neural networks for lvcsr using rectified linear units
  and dropout.
\newblock In {\em ICASSP}, pages 8609--8613. IEEE.

\bibitem[\protect\citename{Ding \bgroup et al.\egroup }2008]{ding2008}
Shilin Ding, Gao Cong, Chin yew Lin, and Xiaoyan Zhu.
\newblock 2008.
\newblock Using conditional random fields to extract contexts and answers of
  questions from online forums.
\newblock In {\em In Proceedings of ACL-08: HLT}.

\bibitem[\protect\citename{Graves}2013]{graves13}
Alex Graves.
\newblock 2013.
\newblock Generating sequences with recurrent neural networks.
\newblock {\em CoRR}, abs/1308.0850.

\bibitem[\protect\citename{Hinton \bgroup et al.\egroup }2012]{hinton2012}
Geoffrey~E. Hinton, Nitish Srivastava, Alex Krizhevsky, Ilya Sutskever, and
  Ruslan Salakhutdinov.
\newblock 2012.
\newblock Improving neural networks by preventing co-adaptation of feature
  detectors.
\newblock {\em CoRR}, abs/1207.0580.

\bibitem[\protect\citename{Hochreiter \bgroup et al.\egroup }2001]{hochreiter}
Sepp Hochreiter, Yoshua Bengio, Paolo Frasconi, and J{\"u}rgen Schmidhuber.
\newblock 2001.
\newblock Gradient flow in recurrent nets: the difficulty of learning long-term
  dependencies.

\bibitem[\protect\citename{Hu \bgroup et al.\egroup }2013]{hu:2013}
Haifeng Hu, Bingquan Liu, Baoxun Wang, Ming Liu, and Xiaolong Wang.
\newblock 2013.
\newblock Multimodal dbn for predicting high-quality answers in cqa portals.
\newblock In {\em Proceedings of the 51st Annual Meeting of the Association for
  Computational Linguistics (Volume 2: Short Papers)}, Sofia, Bulgaria, August.

\bibitem[\protect\citename{Hu \bgroup et al.\egroup }2014]{baotian}
Baotian Hu, Zhengdong Lu, Hang Li, and Qingcai Chen.
\newblock 2014.
\newblock Convolutional neural network architectures for matching natural
  language sentences.
\newblock In {\em Advances in Neural Information Processing Systems 27}, pages
  2042--2050.

\bibitem[\protect\citename{Huang \bgroup et al.\egroup }2007]{huang:2007}
Jizhou Huang, Ming Zhou, and Dan Yang.
\newblock 2007.
\newblock Extracting chatbot knowledge from online discussion forums.
\newblock In {\em Proceedings of the 20th International Joint Conference on
  Artifical Intelligence}, IJCAI'07, pages 423--428.

\bibitem[\protect\citename{Jeon \bgroup et al.\egroup }2005]{jeon:2005}
Jiwoon Jeon, W.~Bruce Croft, and Joon~Ho Lee.
\newblock 2005.
\newblock Finding similar questions in large question and answer archives.
\newblock In {\em Proceedings of the 14th ACM International Conference on
  Information and Knowledge Management}, CIKM '05, pages 84--90.

\bibitem[\protect\citename{Kim}2014]{sentiment}
Yoon Kim.
\newblock 2014.
\newblock Convolutional neural networks for sentence classification.
\newblock {\em CoRR}, abs/1408.5882.

\bibitem[\protect\citename{M{\`a}rquez \bgroup et al.\egroup }2015]{semval}
Llu{\'\i}s M{\`a}rquez, James Glass, Walid Magdy, Alessandro Moschitti, Preslav
  Nakov, and Bilal Randeree.
\newblock 2015.
\newblock Semeval-2015 task 3: Answer selection in community question
  answering.
\newblock In {\em Proceedings of the 9th International Workshop on Semantic
  Evaluation (SemEval-2015)}.

\bibitem[\protect\citename{Mikolov \bgroup et al.\egroup }2013]{wordvec}
Tomas Mikolov, Kai Chen, Greg Corrado, and Jeffrey Dean.
\newblock 2013.
\newblock Efficient estimation of word representations in vector space.
\newblock {\em CoRR}, abs/1301.3781.

\bibitem[\protect\citename{Moschitti \bgroup et al.\egroup
  }2007]{alessandro2007}
Alessandro Moschitti, Silvia Quarteroni, Roberto Basili, and Suresh Manandhar.
\newblock 2007.
\newblock Exploiting syntactic and shallow semantic kernels for question answer
  classification.
\newblock In {\em Proceedings of the 45th Annual Meeting of the Association of
  Computational Linguistics}, pages 776--783.

\bibitem[\protect\citename{Shah and Pomerantz}2010]{Shah:2010}
Chirag Shah and Jefferey Pomerantz.
\newblock 2010.
\newblock Evaluating and predicting answer quality in community qa.
\newblock In {\em Proceedings of the 33rd International ACM SIGIR Conference on
  Research and Development in Information Retrieval}, SIGIR '10, pages
  411--418.

\bibitem[\protect\citename{Srivastava \bgroup et al.\egroup }2015]{lstmuse}
Nitish Srivastava, Elman Mansimov, and Ruslan Salakhutdinov.
\newblock 2015.
\newblock Unsupervised learning of video representations using lstms.
\newblock {\em CoRR}, abs/1502.04681.

\bibitem[\protect\citename{Sutskever \bgroup et al.\egroup }2014]{seq2seq}
Ilya Sutskever, Oriol Vinyals, and Quoc~V. Le.
\newblock 2014.
\newblock Sequence to sequence learning with neural networks.
\newblock {\em CoRR}, abs/1409.3215.

\bibitem[\protect\citename{Wang \bgroup et al.\egroup }2009a]{baoxun2009}
Baoxun Wang, Bingquan Liu, Chengjie Sun, Xiaolong Wang, and Lin Sun.
\newblock 2009a.
\newblock Extracting chinese question-answer pairs from online forums.
\newblock In {\em IEEE International Conference on Systems, Man, and
  Cybernetics (SMC)}, pages 1159--1164.

\bibitem[\protect\citename{Wang \bgroup et al.\egroup }2009b]{wang:2009}
Kai Wang, Zhaoyan Ming, and Tat-Seng Chua.
\newblock 2009b.
\newblock A syntactic tree matching approach to finding similar questions in
  community-based qa services.
\newblock In {\em Proceedings of the 32Nd International ACM SIGIR Conference on
  Research and Development in Information Retrieval}, SIGIR '09, pages
  187--194.

\bibitem[\protect\citename{Wang \bgroup et al.\egroup }2010]{baoxun2010}
Baoxun Wang, Xiaolong Wang, Chengjie Sun, Bingquan Liu, and Lin Sun.
\newblock 2010.
\newblock Modeling semantic relevance for question-answer pairs in web social
  communities.
\newblock In {\em Proceedings of the 48th Annual Meeting of the Association for
  Computational Linguistics}, ACL '10, pages 1230--1238.

\bibitem[\protect\citename{Xue \bgroup et al.\egroup }2008]{xuetrans}
Xiaobing Xue, Jiwoon Jeon, and W.~Bruce Croft.
\newblock 2008.
\newblock Retrieval models for question and answer archives.
\newblock In {\em Proceedings of the 31st Annual International ACM SIGIR
  Conference on Research and Development in Information Retrieval}, SIGIR '08,
  pages 475--482.

\bibitem[\protect\citename{Yu \bgroup et al.\egroup }2014]{dl2qa}
Lei Yu, Karl~Moritz Hermann, Phil Blunsom, and Stephen Pulman.
\newblock 2014.
\newblock Deep learning for answer sentence selection.
\newblock {\em CoRR}, abs/1412.1632.

\bibitem[\protect\citename{Zeiler}2012]{adagrad}
Matthew~D. Zeiler.
\newblock 2012.
\newblock {ADADELTA:} an adaptive learning rate method.
\newblock {\em CoRR}, abs/1212.5701.

\bibitem[\protect\citename{Zhou \bgroup et al.\egroup }2011]{zhou:2011}
Guangyou Zhou, Li~Cai, Jun Zhao, and Kang Liu.
\newblock 2011.
\newblock Phrase-based translation model for question retrieval in community
  question answer archives.
\newblock In {\em Proceedings of the 49th Annual Meeting of the Association for
  Computational Linguistics: Human Language Technologies - Volume 1}, HLT '11,
  pages 653--662.

\end{thebibliography}

\end{document}